%% file: lbpcg_TR.tex
\documentclass[9.5pt,journal]{IEEEtran}
\pdfoutput=1
\usepackage{url}
\usepackage{longtable}
\usepackage{algorithm}
\usepackage{algorithmic}
\usepackage{listings}
\usepackage{graphicx}
\usepackage{subfigure}
\usepackage{pgfplots}
\usepackage{soul}
\usepackage{color}
\usepackage{amsmath}
\usepackage{amssymb}
\usepackage{url}

\newcommand{\bi}{\begin{itemize}}
\newcommand{\ei}{\end{itemize}}
\newcommand{\be}{\begin{enumerate}}
\newcommand{\ee}{\end{enumerate}}
\newcommand{\bc}{\begin{center}}
\newcommand{\ec}{\end{center}}

\newcommand{\beqn}{\begin{eqnarray}}
\newcommand{\eeqn}{\end{eqnarray}}
\newcommand{\beq}{\begin{equation}}
\newcommand{\eeq}{\end{equation}}
\newcommand{\bsubeq}{\begin{subequations}}
\newcommand{\esubeq}{\end{subequations}}
\newcommand{\bcase}{\begin{cases}}
\newcommand{\ecase}{\end{cases}}

\def\BibTeX{{\rm B\kern-.05em{\sc i\kern-.025em b}\kern-.08em
    T\kern-.1667em\lower.7ex\hbox{E}\kern-.125emX}}

\begin{document}

\title{Learning-Based Procedural Content Generation}
\author{Jonathan Roberts and Ke Chen,~\IEEEmembership{Senior Member,~IEEE} %
\IEEEcompsocitemizethanks{\IEEEcompsocthanksitem Authors are with
School of Computer Science, The University of Manchester,
Manchester M13 9PL, United Kingdom.
Email: \{robertsj, chen\}@cs.manchester.ac.uk.}}


\IEEEcompsoctitleabstractindextext{
\begin{abstract}
\input{tex/lbpcg_section_abstract.tex}
\end{abstract}

}

\maketitle

\IEEEdisplaynotcompsoctitleabstractindextext

\section{Introduction}
\label{sec:introduction}
\input{tex/lbpcg_section_introduction.tex}

\section{Background}
\label{sec:background}
\input{tex/lbpcg_section_background.tex}

\section{LBPCG Framework}
\label{sec:lbpcg_framework}
\input{tex/lbpcg_section_framework.tex}

\subsection{Problem Statement}
\input{tex/lbpcg_section_problem.tex}
\subsection{Model Description}
\label{sec:model_description}
\input{tex/lbpcg_section_model_description.tex}

\section{Enabling Techniques}
\label{sec:enabling_techniques}
\input{tex/lbpcg_section_enabling.tex}

\subsection{ICQ Learning}
\input{tex/lbpcg_section_icq.tex}

\subsection{Learning for CC}

\input{tex/lbpcg_section_cc.tex}

\subsection{GPE Learning}

\input{tex/lbpcg_section_gpe.tex}

\subsection{Learning for PDC}

\input{tex/lbpcg_section_pdc.tex}

\subsection{On-line Generation with the IP Model}
\input{tex/lbpcg_section_ip.tex}

\section{LBPCG-Quake: A Proof of Concept}
\label{sec:lbpcg-quake}
\input{tex/lbpcg_section_quake.tex}

\subsection{Overview}
\input{tex/lbpcg_section_quake_overview.tex}

\subsection{Data Collection}
\label{subsect:quake_data_collection}
\input{tex/lbpcg_section_data_collection.tex}

\subsection{LBPCG-Quake Learning}
\label{subsect:quake_learning}
\input{tex/lbpcg_section_lbpcg_quake_learning.tex}

\subsection{Simulation Results}
\input{tex/lbpcg_section_simulation_result.tex}

\section{Discussion}
\label{sec:discussion}
\input{tex/lbpcg_section_discussion.tex}

\section*{Acknowledgement}
\input{tex/lbpcg_section_acknow.tex}

\end{document}

%% file: tex/lbpcg_section_abstract.tex
Procedural content generation (PCG) has recently
become one of the hottest topics in computational intelligence and
AI game researches. Among a variety of PCG techniques, search-based
approaches overwhelmingly dominate PCG development at
present. While SBPCG leads to promising results and
successful applications, it poses a number of challenges ranging
from representation to evaluation of the content being generated.
In this paper, we present an alternative yet generic PCG framework, named
\emph{learning-based procedure content generation} (LBPCG),
to provide potential solutions to several
challenging problems in existing PCG techniques.
By exploring and exploiting information gained in game development
and public beta test via data-driven learning,
our framework can generate robust content adaptable to
end-user or target players on-line with minimal interruption to their experience.
Furthermore, we develop enabling techniques to implement the various models required
in our framework. For a proof of concept, we have developed a
prototype based on the classic open source first-person shooter game, Quake.
Simulation results suggest that our framework is promising in
generating quality content.

%% file: tex/lbpcg_section_introduction.tex
The video games industry has been expanding rapidly and even surpassed the movie industry in revenue \cite{urlBiggerThanMovies}. The expectations of consumers have gradually increased to the point where players and critics demand cutting edge graphics, immersive game play and strong replay value from new releases. Generating game content is costly, as it requires many different skill sets and long periods of development. For example, generating the skeletal animation for a single character can take several months. It is therefore of the upmost importance to create content in timely and efficient manner.
\par

Procedural Content Generation (PCG) is the process of generating content for a video game automatically using algorithms. A wide variety of content can be generated, e.g., art assets such as terrain and textures and even high level game play structures such as storyline and characters. PCG not only has the potential to provide a basis built upon by developers but also can provide an endless stream of content for a player to extend the lifetime of the game. If used properly, it can reduce the amount of resources required to build a game and thus drive down costs. Although PCG often seems like a good approach, it may be very tricky to get right and could actually be more trouble than creating handcrafted content \cite{urlProceduralHard}.
\par

In recent years, a variety of approaches have been proposed to improve PCG. Most of such approaches generally fall under the umbrella of \emph{search-based procedural content generation} (SBPCG), a terminology coined by Togelis et al \cite{sbpcg}. Given a potential game content space, SBPCG employs evolutionary or other metaheuristic search algorithms to explore the space efficiently to find content appealing to players. This can result in more robust and trustworthy PCG algorithms that require far less manual configuration than traditional approaches. In the past few years, researchers have used different SBPCG techniques to adapt levels for games such as 2D platformers \cite{Laskov2009,Shaker2010Towards,Shaker2012EvolvingPersonalized}, first-person shooters \cite{Cardamone2011Evolving}, rogue-like games \cite{Togelius2012Compositional}, racing games \cite{Togelius2007} and real-time strategy games \cite{Liapis2012Limitations}. Some of the latest work in SBPCG was collected in a special issue on PCG \cite{special-issue11}.
\par

While SBPCG leads to promising results and successful applications, it also poses a number of challenges as pointed out in \cite{sbpcg}. As a special case of the generate-and-test PCG approach, SBPCG needs to address three critical issues \cite{sbpcg}: \emph{content representation},
\emph{search space management} and \emph{content test} with evaluation functions. As a widely used technique in SBPCG, evolutionary computation turns out to be effective
due to its nature but encounters difficulties in tackling problems regarding the three aforementioned issues. First of all, an evolutionary algorithm has to deal with the genotype-to-phenotype mapping. Although there are direct encodings, indirect encodings are often used in games for computational efficiency but they may be short of a clear genotype-phenotype distinction. Moreover, it is generally not guaranteed that such representations always preserve the locality property \cite{sbpcg}, which may result in ineffective search and even entail a risk of catastrophic failure. Next, many games have a large content space and SBPCG may take a long time to generate quality content. Thus, SBPCG seems more suitable for off-line other than online content generation. Finally, a population of content needs to be generated and evaluated via the guidance of various fitness or evaluation functions. Fitness functions may be ill-posed and subjective, which causes difficulties in developing the accurate fitness functions encoding developers' prior knowledge and players' experiences \cite{experience11}.
\par

In this paper, we propose a novel framework, termed the learning based procedural content generation (LBPCG), which aims to address the main issues encountered by existing PCG techniques in the context of commercial video game development.
Unlike SBPCG, the LBPCG gains information and encoding knowledge/experience via data-driven machine learning from different contributors involved in a typical video game development life-cycle, such as developers and public beta testers but minimizes interference in the end-users experience.
Relying on generalization of learning-based component models, our framework acts as a controller for game content generation and tends to create ideal content for an arbitrary player. As a result,
the LBPCG has several considerable advantages including avoiding hard coded evaluative functions, flexibly matching between content and players' types/experience for adaptivity, and not interfering with the players' experience.
\par

 In this paper, our main contributions are summarized as follows: a) we propose a novel learning-based PCG framework of the data-driven nature to overcome weaknesses over existing PCG techniques, b) we develop enabling techniques to support the proposed LBPCG framework, and c) we apply the LBPCG to a classic first-person shooter game, \emph{Quake} \cite{Quake}, for a proof of concept, which leads to a prototype generating quality content as verified in our simulations.
\par

 The rest of paper is organized as follows. Sect. \ref{sec:background} reviews relevant background.  Sect. \ref{sec:lbpcg_framework} presents the LBPCG framework. Sect. \ref{sec:enabling_techniques} describes enabling techniques to support the LBPCG implementation. Sect. \ref{sec:lbpcg-quake} describes a proof of concept by applying the LBPCG framework to Quake. The last section discusses issues related to the LBPCG framework.

%% file: tex/lbpcg_section_background.tex
In this section, we review the main concerns over existing PCG techniques in the context of commercial video game development.
\par

Commercial video games development is usually very well-structured, incorporating many work-flows for design, testing and deployment. Most companies have a cohort of personnel consisting of programmers, designers, artists, production managers and testers. In this paper, we term such people \emph{developers}. Games will go through many iterations, where features are added, tested by engineers, then released for the test teams. After a game has reached a certain level of maturity, it is sometimes released to the public as a beta test (hereinafter all participants are dubbed \emph{beta testers}).  Besides bug finding, the other reason behind doing this is that in the games industry it is common for developers to be happy with the game they have created, but public opinion can be very different. Garnering public opinion is useful and can avoid the disastrous event of releasing a bad, expensively made, game to the public. After all production tests pass, the game goes ``gold" and is released to the publishers for distribution to the end-user, \emph{target players} in our terminology.
\par

 In this paper, we confine ourselves to a class of \emph{content generators} that create expanded content based on a parsimonious yet generic representation, i.e., a parameterized \emph{content description vector} that specifies the degree of control of content to be generated along with an uncontrollable random seed.   An example of such a content generator would be a level generator, which takes a multidimensional parameter vector saying how many monsters, weapons, health, ammunition, doors and puzzles there will be in the generated level.  For a content generator, its variable parameters generally define a \emph{content space} of all possible content that can be generated by this generator. In general, such a content space is likely to be large enough such that exhaustive exploration of all content before-hand is infeasible.
 We believe that the ultimate goal is to manipulate the parameters for a content generator to build the best content for a target player's individual tastes.
 To attain this goal, however, existing PCG techniques, including SBPCG, encounter a number of challenges \cite{sbpcg}.
\par

First of all, the risk of catastrophic failure and a lack of quality become a main concern for PCG given the fact that a content generator may generate unplayable content or content that is not appropriate for anyone's tastes. Illegitimate content has some extremely undesirable property such as being unbeatable, e.g., a map that consists of rooms full of monsters that immediately kill the player no matter their skill level. Some efforts were made to ward off this problem.
For example, studies in \cite{Togelius2012Compositional,Shaker2012EvolvingPersonalized} describe methods by combining off-line and online models, one for optimizing playability/winnability and the other for optimizing challenge/skill, with some success. The integrated model may avoid catastrophic failure and then adapt the generated content to suit a specific player online. Although some progress was made in the experiment with Infinite Mario \cite{Shaker2012EvolvingPersonalized}, it is still unclear to what extent models of different functionality and purpose can be combined to form robust composite models.
\par

Next, personalization of content for a target player is driven by not only factors such as playability and skills but also a player's affective and cognitive experience \cite{experience11}, which poses another big challenge to PCG.
To generate the favored content for a target player,
player styles/types need to be identified correctly and their experience must be measured accurately. Obviously, it is undesirable to interfere with the target player's experience, so asking them any questions such as ``how much fun was that content?" should be avoided.
To our knowledge, however, most of existing methods need to identify the target player's style/type and to measure their experience by learning from their feedback \cite{Yannakakis2009,Pedersen2010} or behavior \cite{experience11}, which leads to a burden to target players.
Moreover, existing methods often utilize pre-defined player types/styles, e.g., traits or personalities in terms of psychology, to categorize target players explicitly, which might not encapsulate all player types and lead to difficulty to infer players' types/styles as pre-defined. Furthermore, players' subjective feedback could be quite noisy and inaccurate, which exacerbates difficulties.
\par

To avoid modeling target players directly, it is possible to learn a generic playing style/type model from beta testers feedback and behavior \cite{experience11,LDA12,Mahlmann2010Predicting}.
Nevertheless, data collection in such methods often undergoes an expensive and laborious process, as collecting data from members of the public often intuitively seems like the best way to build up realistic models of enjoyment in humans. Recently, a system was built up in order to predict human behavior based on game-play statistics produced via a pool of 10,000 players \cite{Mahlmann2010Predicting}. Their study concluded that one of challenges faced in such work was the existence of many outliers, people who behave in an irregular fashion \cite{Mahlmann2010Predicting}. If one was to collect direct feedback from the public with questions, e.g., by asking them whether they were having fun, it would be expected that many people would be difficult to process not only because they were outliers, but also because they may deliberately or accidentally provide feedback that does not match their true opinions. When crowd-sourcing is applied to PCG, how to deal with noisy data therefore is an ongoing critical issue.
\par

Finally, one has to be mindful that a players' notion of fun (or whatever other attribute is being optimized) may drift over time when adapting content to a target player for personalization. In the work with creating an adaptive Super Mario \cite{Shaker2010Towards}, such an issue was touched upon somewhat in their experiment. On the one hand, their system was fairly successful in a test by switching the agent that content is being optimized for half-way through. With the same experimental protocol on human players, however, it was reported that only 60\% of players preferred their dedicated system over a purely random system \cite{Shaker2010Towards}. It is therefore apparent that another open problem in PCG is properly safeguarding online algorithms against players who may change their mind over time, which is perfectly reasonable in the case of adapting an attribute such as difficulty.
\par

%% file: tex/lbpcg_section_framework.tex
In this section, we first describe our problem statements resulting from the main concerns reviewed in last section and present a generic PCG framework to address all the issues in the context of commercial video game development.
\par

%% file: tex/lbpcg_section_problem.tex
By taking the main concerns over existing PCG techniques into account in the context of commercial
video game development, we list a number of emerging problems in PCG as follows:
\be
\item how to encode developers knowledge on the content space represented via game parameters to avoid completely undesirable content by limiting search space effectively,
\item how to enable developers to define controllable player type/style in an implicit yet flexible way for player categorization,
\item how to learn robust and ``true" players' experience via crowd-sourcing by filtering out outliers and untrustworthy feedback,
\item how to exploit the information from the beta testers experience gained via crowd-sourcing to categorize players in terms of player type/style specified by developers,
\item how to tackle the concept-drift problem for target players, due to the change of players' experience, faced by online adaptive algorithms.
\ee
\par
While the aforementioned problems may be studied separately, we propose a generic PCG framework to tackle all the issues systematically.

%% file: tex/lbpcg_section_model_description.tex
Motivated by the commercial video game development process, we would divide a typical game life cycle into three stages: \emph{development} (involving developers), \emph{public test} (involving beta testers) and \emph{on-line} (concerning target players). We believe that a PCG approach should make the best use of developers knowledge during development and the information gained from beta players in public test so that it can generate tailored content
 at the on-line stage for target players while minimizing interruptions to them.
 To attain such a goal, we propose a \emph{Learning-Based Procedural Content Generation} (LBPCG) framework. The general ideas behind our LBPCG framework are two-fold: i) encoding developers knowledge on content and modeling public players' behavior/experience via machine learning, ii) strategically using all learned models to control a content generator to produce on-line content that ``matches" target players' individual preference via prediction based on their behavior. As depicted in Fig. 1, our LBPCG framework consists of five models: \emph{Initial Content Quality} (ICQ), \emph{Content Categorization} (CC), \emph{Generic Player Experience} (GPE), \emph{Play-log Driven Categorization } (PDC) and \emph{Individual Preference} (IP). The ICQ and the CC models encode developers knowledge on content to tackle the first two issues, while the GPE and the PDC models are used to model beta players' behavior/experience and to link those to developers knowledge on content, which deals with the third and the fourth issues. By means of all other four models, the IP model is designed to generate quality content on-line for target players where the last issue is addressed.
\par

\begin{figure}[t]
\bc
\includegraphics[scale=0.22]{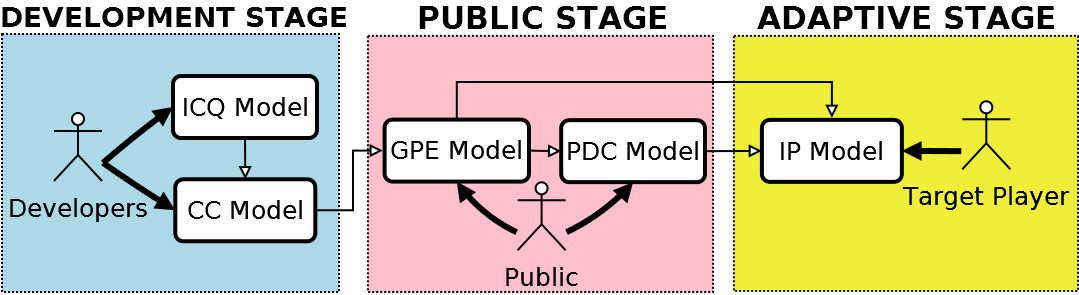}
\ec
\label{fig:lbpcg}
\vspace*{-3mm}
\caption{Learning-based procedure content generation (LBPCG) framework.}
\vspace*{-5mm}
\end{figure}

The ICQ model is used to filter out \emph{illegitimate} and/or \emph{unacceptable} content
of poor quality. One can view the ICQ model as carving out a manifold of acceptable content in the original content vector space. It is from within this manifold that the framework will eventually select content for a target player, which is expected to limit the content search space significantly.
As content is represented by a game parameter vector, the task in the ICQ model
appears to be a binary classification problem; i.e., for a parameter vector, the ICQ model decides if
it leads to acceptable or unacceptable content. While this problem might be tackled in different ways,
we prefer the learning-based methodology since it encodes developers knowledge in an implicit way and avoids handcrafted work having to be done by developers. An ideal scenario would be as follows. By playing a few well-selected games, developers annotate them as either acceptable or unacceptable based on their knowledge. For annotated games, their parameter vectors and developer assigned labels form training examples to train an ICQ classifier. Then the classifier would be able to make a correct decision for any game parameter vector. In order to build up such a learning-based ICQ model, there are two critical issues: how to select a ``minimal" number of representative games for annotation and how to build up a learner of good generalization.
\par

Our motivation underlying the CC model comes from our observations that one prefers content with similar game features to games they have already enjoyed. Thus, preference for content features can be
utilized in specifying players' styles/types in an implicit yet flexible way. The CC model is designed to categorize all acceptable content in terms of content features defined by developers. By using such content features, acceptable content of similar properties would be grouped together to form game categories identified with a unique index. Content features may be generic, e.g., challenge, frustration, and fear induction or specific to a game in hand, e.g., categorical amounts of weapons/monsters in first person shooter games. Depending on the content feature nature, the problem of the CC model may be either regression or classification. For the same reasons as described above for the ICQ, once again, we advocate the data-driven methodology. Consequently, annotation needs to be performed by developers for each content feature, and the same issues in the ICQ learning need to be addressed for a learning-based CC model. For learning the ICQ and the CC models, the only difference is that the ICQ model works on the whole content space while the CC model needs to only take surviving content subspaces (those passing the ICQ examination) into account and annotated games used in the ICQ learning may be reused to train the CC model.
\par

It is well known that public beta tests may provide valuable information for game development.
In our framework, we explore and exploit such information via well-selected representative games and an
elaborately designed survey. Ideally, developers have sufficient knowledge on content space after
the ICQ and the CC models are established. By means of these two models, developers should be able to select
a small number of representative games of quality content from each game category identified with
content features. Instead of releasing all acceptable games to beta testers, we suggest releasing only well-selected games in an initial beta test. The motivation behind this suggestion is that we would use data including play-logs and feedback to facilitate modeling players' styles/types pre-defined by developers when the CC model was built up. As a result, the GPE model is proposed to undertake this task. As reviewed in Sect. II.A, public opinions may be different from developers' and their feedback is often subjective and very noisy. Hence the GPE model needs to address two critical issues: how to find a ``genuine" consensus on each selected game without being compromised by deceitful or deliberately inaccurate feedback and how to assign a confidence level on reliability to each beta player who provides feedback. By dealing with two issues properly, the GPE model would rank all selected games in terms of public consensus and decide the reliability of all participated beta testers. It is also worth mentioning that the GPE model can be used to deliver a stream of games that players are probably going to enjoy to some extent. Hence, such games should be used in initial individual preference detection in the IP model to minimize interruption to players' experience. In the absence of an accurate personalized model for a player, the GPE model further provides a decent back-up option, which will be clarified in the description of the IP model below and its enabling technique in the next section.
\par

As the GPE model provides the reliability information of all beta testers, those unreliable beta testers can be
 easily thrown out. For reliable beta testers, their play-logs contain rich information in diversified players' behavior and the feedback to games they played provides information on their ``genuine" preferences of the specific content features pre-defined by developers in the CC model. By associating the play-logs on a specific game and its content features with corresponding feedback, it is straightforward to establish the direct correspondence between players' behavior recorded in play-logs and preferred games characterized by content features pre-defined by developers. Thus, such correspondence would be able to categorize players based on their behavior on specific games. In the LBPCG framework, we employ the PDC model to take on this task, and building up the PDC model can be formulated as a supervised learning problem. For learning, a training example takes both a play-log on a specific game and its content features as the input and the corresponding feedback as the target. Once the PDC model is trained, it should be able to predict preferences of target players via their play-logs and pre-defined content features of games played. By working with the CC model seamlessly, the PDC model yields a novel measure to categorize target players implicitly, which tends to minimize interrupting target players' experience.

After the above four models are accomplished, the IP model is required at the online stage to generate the ideal content for a target player by applying all four models strategically. An ideal scenario for the IP model is as follows. After a target player has played few well-selected games based on the ICQ, the CC and the GPE models, his/her categorical preference is determined rapidly via the PDC model. Then quality games with variation in his/her preferred content category are generated subsequently until the concept-drift problem occurs. Once such a problem is detected, the new categorical preference of the target player is quickly determined by repeating the initial detection process so that his/her favorite content in a new category can be generated.  To enact such a scenario, the IP model needs to cope with three critical issues: how to find a ``minimal" number of appropriate games initially to ensure the categorical preference of a target player can be determined rapidly, how to ensure sufficient content variation in subsequent generated games within a particular content category once the player type is determined and how to detect when concept-drift occurs and tackle it effectively.
\par

In summary, our LBPCG framework along with the enabling techniques presented in the next section provides a systematic solution to emerging problems over existing PCG techniques. Here, we emphasize that the use of data-driven learning has distinct advantages in encoding developers knowledge and modeling beta players' behavior/experience, which will be further discussed in  Sect. \ref{sec:discussion}.

%% file: tex/lbpcg_section_enabling.tex
In this section, we present enabling techniques used to support our LBPCG framework. For each model in LBPCG, we first formulate problems and then describe our solutions.
\par

%% file: tex/lbpcg_section_icq.tex
The goal of the ICQ model is to be able to recognize whether a content description vector represents an acceptable game or not. In general, we denote a content space of $D$ parameters as $\mathcal{G} \!\subset\! \mathbb{R}^D$ where
${\boldsymbol g} \!\in\! \mathcal{G}$ is the parameter vector of a game
denoted by ${\boldsymbol g} \!=\! (g_1, g_2, \cdots, g_D)$.
As such, the problem in the ICQ model is finding out a mapping for any ${\boldsymbol g} \in \mathcal{G}$ so that
$\Phi_{I\!C\!Q}\!: {\boldsymbol g}\! \rightarrow \!\{+1, -1\}$ where $+1$/$-1$ indicates the parameter vector, ${\boldsymbol g}$, leads to acceptable/unacceptable content, respectively.
\par

\begin{algorithm}[t]
\caption{\bf Active ICQ Learning}
\label{alg:icq_active}
\begin{algorithmic}
\STATE{\textbf{Initialization}: annotate a few randomly selected games beyond $T_{I\!C\!Q}$
and then pre-train a chosen binary classifier with these training examples}
\WHILE{test error on $T_{I\!C\!Q}$ is not acceptable}
\FORALL{${\boldsymbol g} \in \mathcal{G}'$ }
	\STATE Test ${\boldsymbol g}$ with the classifier of the current parameters
	\STATE Record its confidence on the label assigned to ${\boldsymbol g}$
\ENDFOR
\STATE Find the game of the least confidence, ${\boldsymbol g}^*$
\STATE Annotate ${\boldsymbol g}^*$ by developers if it was not annotated
\STATE Update parameters of the classifier with ${\boldsymbol g}^*$ and its true label
\ENDWHILE
\end{algorithmic}
\end{algorithm}

 In general, content space is often too large to be exhaustively classified. To tackle this problem, we employ a  binary classifier of confidence on its decision \cite{duda01}  by training it on a number of annotated games by developers.  As the developer is a limited resource, we want to keep this number to as few as possible, while being assured we are sampling from a good spectrum of games. Instead of randomly selecting games for annotation, we suggest applying a clustering analysis algorithm \cite{duda01}
  to $\mathcal{G}$, which leads to a partition of $K$ clusters. Then, we choose a fixed number of games from each cluster, which results in much smaller search space, $\mathcal{G}'$, that is spread throughout the entire original space, $\mathcal{G}$, without being condensed in one particular area. By avoiding a condensed area, we can be somewhat assured that we will not be learning from games that are extremely similar.
 \par

 For annotation of a game, developers need to play the game and the assign a legitimacy label to it based on their knowledge. Apart from this label, developers also assign labels/scores on content features pre-defined by themselves, which can be used in the CC model described later on. Consolidating data annotation/collection in this way further reduces the workload on developers.

\par

 While there are different methods to train a classifier, we propose a query-based active learning method especially for ICQ learning.  Active learning aims to reduce the burden on developers by asking them as few questions as possible. Our algorithm searches for training examples of the least confidence, asks developers to annotate only games of the least confidence and then re-trains the classifier until convergence. Before our algorithm is applied, developers need to annotate at least a representative game from each of $K$ clusters archived by clustering analysis. All the annotated games constitute a validation set, $T_{I\!C\!Q}$. As a result,
 our active learning algorithm is summarized in Algorithm \ref{alg:icq_active}.
\par

By applying this active learning model, the whole content space, $\mathcal{G}$, is reduced to a smaller search space, $\mathcal{G}_a$,  that tends to contain only acceptable content. Instead of searching $\mathcal{G}$, all subsequent models will work on $\mathcal{G}_a$ only.

%% file: tex/lbpcg_section_cc.tex
The purpose of the CC Model is to categorize acceptable content based on $F~(F \ge 1)$ pre-defined content features denoted by ${\boldsymbol c}=(c_1,\cdots,c_F)$ where a feature $c_f$ may have a continuous or discrete value. As such the problem in the CC model is finding out a mapping for ${\boldsymbol g} \in \mathcal{G}_a$ so that
$\Phi_{C\!C}\!: {\boldsymbol g}\! \rightarrow \! c_f$ for $f\!=\! 1,\cdots,F$ where the value of $c_f$ indicates its
characteristic or category of ${\boldsymbol g}$ in terms of content feature $c_f$. Depending on the
nature of content features, we can formulate this problem as either regression or multi-class classification.
\par

\begin{algorithm}[t]
\caption{\bf Active Learning for CC}
\label{alg:cc_active}
\begin{algorithmic}
\STATE{\textbf{Initialization}: annotate a few randomly selected games beyond $T_{C\!C}$
and then pre-train $F$ chosen learners, respectively, with these training examples}
\FORALL{$c_f \in {\boldsymbol c}$}
\WHILE{test error on $T_{C\!C}$ is unacceptable in terms of $c_f$}
\FORALL{${\boldsymbol g} \in \mathcal{G}_a'$ }
	\STATE Test ${\boldsymbol g}$ with the $f$th current learner
	\STATE Record its confidence on the label assigned to ${\boldsymbol g}$
\ENDFOR
\STATE Find the game of the least confidence, ${\boldsymbol g}^*$
\STATE Annotate ${\boldsymbol g}^*$ by developers if it was not annotated
\STATE Update parameters of the $f$th learner with ${\boldsymbol g}^*$ and its ground-truth
\ENDWHILE
\ENDFOR
\end{algorithmic}
\end{algorithm}

As described in Sect. III.B, there are similar issues to those arising in the ICQ model; the acceptable content
space, $\mathcal{G}_a$, is generally still too large to be exhaustively searched and the number of annotated games need to be as few as possible, while being assured sampling is from a good spectrum of games in $\mathcal{G}_a$.
Therefore, we proceed in a similar fashion as with ICQ learning by sharing resource and exploiting results achieved in ICQ learning, including annotated data and the clustering analysis results. With the same motivation as is in
ICQ learning, we propose a generalized active learning algorithm for CC. By using the same
partition of $K$ clusters achieved in ICQ learning, we choose the fixed number of acceptable games in $\mathcal{G}_a$ from each cluster, which results in much smaller search space, $\mathcal{G}_a'$, that is spread throughout the entire acceptable content space, $\mathcal{G}_a$, without being condensed in one particular area. Furthermore,
developers need to ensure there is at least a representative acceptable game to be annotated from each of $K$ clusters by re-using all acceptable games in $T_{I\!C\!Q}$ and annotating some games if no annotated games in a cluster are available. All the annotated acceptable games constitute a validation set, $T_{C\!C}$. The generalized active learning algorithm for CC is summarized in Algorithm \ref{alg:cc_active}.
\par

After active learning finishes for the CC, $F$ resultant learners are ready to predict content features for any acceptable games in $\mathcal{G}_a$. For each content feature, $c_f$, the $f$th learner outputs a predicted label/value and confidence on such a prediction.  Note that incorrectly characterized games will corrupt our individual preference detection process as is applied in the IP model. To alleviate the problem, we can use confidence to reject games so that only the acceptable content subspace of high confidence,  $\mathcal{G}_{a\!c}$,  be used at the on-line stage for content generation. For the same reason, we emphasize that it is necessary in learning for CC to have sufficient training examples for each content feature, otherwise variation would be reduced. Thus, content features and their value domain should be pre-defined and selected properly by developers to avoid difficulties and over-complexity in learning for CC.

%% file: tex/lbpcg_section_gpe.tex
The role of the GPE model is two-fold: modeling the public consensus regarding the highly confident acceptable content subspace, $\mathcal{G}_{a\!c}$, and identifying reliable beta testers.
For doing so, developers need to select $N$ quality games, $G=\{{\boldsymbol g}_1, {\boldsymbol g}_2, \cdots, {\boldsymbol g}_N\}$, that span the full categorical spectrum
defined by content features, which can be done by applying developers' knowledge/experience with the ICQ and the CC models and reusing annotated games achieved in learning for CC. These selected games are presented to $P$ beta players, from whom their feedback regarding their experience, e.g., fun or not fun, will be collected and their play-logs will be recorded for any games they played. In general, the feedback is often characterized by a multi-valued ordinal variable, $y$, e.g., how fun on a scale of 1 to 5. For a game, ${\boldsymbol g}_n$ played by player $p$, his/her feedback in the survey is denoted by $y_n^{(p)}$.
\par

The problem in the GPE model is formulated as follows: for any game ${\boldsymbol g}_n \in G$ played by beta players, finding the consensus experience $\hat{y}_n$ on this game with a confidence level $\gamma_n$ and assigning a reliability factor to each beta player, $p$, with his/her feedback comparing to other players' after he/she played a number of games in $G$. Note that multi-valued feedback can be viewed as multiple binary feedback. To simplify the presentation, we only take the binary feedback into account. As such $ y \! \in \! \{1,0\}$ where $1/0$ indicates positive/negative feedback, respectively, and the reliability factor is specified via the \emph{sensitivity} (true positive rate), $\alpha^{(p)}$, and the \emph{specificity} (true negative rate), $\beta^{(p)}$. This is a typical crowd-sourcing problem that seeks a solution to inferring a ``true" label for an object from noisy labels given by multiple annotators. While there are many potential solutions to such a problem, we adapt a generic crowd-sourcing algorithm \cite{Crowd_EM}, hereinafter named \emph{Crowd-EM}, for our enabling technique as it meets our general requirements and can be extended to any form of feedback including continuous and categorical scores \cite{Crowd_EM}. The crowd-EM algorithm for learning GPE is summarized in Algorithm \ref{alg:crowd_em}. Note that a player might not play all
$N$ games. In this case, one simply needs to substitute $P$, the number of all beta players,  with  $P_n$, the number of beta players who actually played game ${\boldsymbol g}_n$, as well as $N$, the number of all games in $G$, with $N_p$, the number of games that player $p$ actually played, in Algorithm \ref{alg:crowd_em}.
\par
\begin{algorithm}[t]
\caption{\bf Crowd-EM for Learning GPE}
\label{alg:crowd_em}
\begin{algorithmic}
\STATE{\textbf{Initialization}\\}
\STATE{For $p\!=\! 1,\cdots,P$, set $\alpha^{(p)}(0)=\beta^{(p)}(0)=0.5$.}
\STATE{For $n\!=\! 1,\cdots,N$, set $\gamma_n(0)=\frac 1 P \sum_{p=1}^P y_n^{(p)}$.}
\STATE{Pre-train a chosen regressor,
$f({\boldsymbol g}, \Theta)$, by finding optimal parameters, $\Theta^*(0)$,
with the training set of $N$ examples,
$\big \{ \big ({\boldsymbol g}_n, \mu_n(0)\big ) \big \}_{n=1}^N$, and set $t\!=\!1$.
\vspace*{2mm}}
\STATE{\textbf{E-Step}\\}
\STATE{For $n\!=\! 1,\cdots,N$, calculate
\vspace*{-2mm}
$$
h_n(t)=f({\boldsymbol g}_n,\Theta^*(t\!-\!1)),
$$
\vspace*{-2mm}
$$
a_n(t) = \prod_{p=1}^P \big [ \alpha^{(p)}(t\!-\!1) \big]^{y_n^{(p)}}
\big [ 1- \alpha^{(p)}(t\!-\!1) \big]^{1-y_n^{(p)}},
$$
\vspace*{-2mm}
$$
b_n(t) = \prod_{p=1}^P \big [ \beta^{(p)}(t\!-\!1) \big]^{1-y_n^{(p)}}
\big [ 1- \beta^{(p)}(t\!-\!1) \big]^{y_n^{(p)}},
$$
$$
\gamma_n(t)= \frac {a_n(t) h_n(t)} {a_n(t) h_n(t) + b_n(t)[1- h_n(t)]}.
$$
}
\STATE{\textbf{M-Step}\\}
\STATE{For $p\!=\! 1,\cdots,P$, update
\vspace*{-2mm}
$$
\alpha^{(p)}(t)=\frac {\sum_{n=1}^N\gamma_n(t) y_n^{(p)}} {\sum_{n=1}^N\gamma_n(t)},
$$
\vspace*{-2mm}
$$
\beta^{(p)}(t)=\frac {\sum_{n=1}^N[1-\gamma_n(t)](1- y_n^{(p)})}
{\sum_{n=1}^N[1-\gamma_n(t)]}.
$$
\vspace*{-2mm}
}
\STATE{Re-train the chosen regressor,
$f({\boldsymbol g}, \Theta)$, by finding optimal parameters, $\Theta^*(t)$,
with the training set of $N$ examples,
$\big \{ \big ({\boldsymbol g}_n, \mu_n(t)\big ) \big \}_{n=1}^N$, and set $t\!=\!t\!+\!1$.
\vspace*{2mm}}
\STATE{\textbf{Repeat} both \emph{E-Step} and \emph{M-Step} \textbf{until} convergence.}
\end{algorithmic}
\end{algorithm}
\par

After the GPE learning is accomplished, the model itself will provide two pieces of important information as follows. Firstly, the GPE model assigns the reliability factor $(\alpha^{(p)}, \beta^{(p)})$ to each of $P$ beta players
who participated in the survey. Secondly, a positive consensus experience score for an acceptable game ${\boldsymbol g} \in \mathcal{G}_{a\!c}$ can be predicted via the trained regressor $\gamma = f({\boldsymbol g}, \Theta^*)$ where $\gamma$ indicates the confidence for such a score.  The former will be utilized in learning for PDC to get rid of unreliable beta players or outliers in order to ensure that ``genuine" players' behavior and experience are learned.
The latter will be exploited in the IP model to facilitate individual preference detection when there is little information on a target player and to serve as a decent fall-back solution when a target player is difficult to categorize.

%% file: tex/lbpcg_section_pdc.tex
The goal of the PDC model is accurately mapping target players' behavior on a game they played and
the game category information characterized by its content features onto their individual preference.
To create such a model, we explore and exploit information provided by beta players in their survey.
In general, we denote a play-log of $L$ event attributes as $\mathcal{L} \!\subset\! \mathbb{R}^L$ where ${\boldsymbol l} \!\in\! \mathcal{L}$ is the play-log attribute vector,
denoted by ${\boldsymbol l} \!=\! (l_1, l_2, \cdots, l_L)$, recorded during play
a game (generated via its parameter vector, ${\boldsymbol g}$). With the same notation used before,
the problem in the PDC model is finding out a mapping for
$({\boldsymbol l}, {\boldsymbol c})$ so that
$\Phi_{P\!D\!C}\!: ({\boldsymbol l}, {\boldsymbol c})\! \rightarrow \! y$.
Thus, we can formulate this problem as multi-class classification in general.
\par
During public beta test, $N_T$ play-logs along with their corresponding feedback are recorded by a data collection system. By combining with content features of each of the played games, we
have a training data set of $N_T$ examples, $\mathcal{D} \!=\! \big \{ [({\boldsymbol l}_i, {\boldsymbol c}_i), y_i] \big \}_{i=1}^{N_T}$. However, such examples may be quite noisy as previously discussed. As a result, we explore the reliability information of each beta player,  $(\alpha^{(p)}, \beta^{(p)})$, learnt by the GPE model to filter out noisy examples. By setting thresholds for  $\alpha^{(p)}$ and $\beta^{(p)}$, respectively, we can generate a  training subset of examples offered by only beta players whose $\alpha^{(p)}$ and $\beta^{(p)}$ are above thresholds. Threshold setting is a challenging problem since there is no clear criterion for it and a low reliability score assigned to a player does not have to mean that the player is a ``liar" but may suggest he/she be an ``outlier" who has distinct experience from the majority of beta players. To tackle this problem, we propose an ensemble learning algorithm for the PDC learning. The basic idea is using different thresholds to produce $M~(M\!>\!1)$ training sets and then train $M$ classifiers on those training data subsets via cross-validation, respectively. Accuracy measured on cross-validation would be used as weights to construct an ensemble classifier. Our ensemble learning algorithm is as summarized in Algorithm \ref{alg:pdc_ensemble}.

\par

\begin{algorithm}[t]
\caption{Learning for PDC}
\label{alg:pdc_ensemble}
\begin{algorithmic}
\STATE{\textbf{Initialization}:
Divide the training data set, $\mathcal{D}$, collected in beta tests into $M$ training subsets, $\mathcal{D}_1,\cdots,\mathcal{D}_M$ by setting
 a number of thresholds for $\alpha^{(p)}$ and $\beta^{(p)}$, respectively. Choose a proper classifier, $f[({\boldsymbol l}, {\boldsymbol c}),\Theta]$.}
\FOR{$m = 1$ \TO $m = M$}
\STATE {Train $f[({\boldsymbol l}, {\boldsymbol c},\Theta_m)]$ on $\mathcal{D}_m$ via cross-validation
by finding optimal parameter, $\Theta_m^*$.}
\STATE {Record its accuracy, $u_m$, on cross-validation.}
\ENDFOR
\STATE{Calculate weights:
\vspace*{-2mm}
$$w_m=\frac {\exp(u_m)} {\sum_{k=1}^M \exp(u_k)} ~~{\rm for}~  m\!=\!1,\cdots,M.$$
\vspace*{-2mm}
}
\STATE{Construct the ensemble classifier:
\vspace*{-2mm}
$$F[({\boldsymbol l}, {\boldsymbol c}),\Theta^*]=\sum_{k=1}^M w_k
f[({\boldsymbol l}, {\boldsymbol c}),\Theta_k^*].$$}
\vspace*{-3mm}
\end{algorithmic}
\end{algorithm}

\par

Once the PDC model is built up with Algorithm \ref{alg:pdc_ensemble}, it will be used at the on-line stage for the IP model to predict individual preference of target players based on content features of games they played and their corresponding play-logs.

%% file: tex/lbpcg_section_ip.tex
\begin{figure}[t]
\label{fig:ip_categorize}
\bc
\includegraphics[scale=0.26]{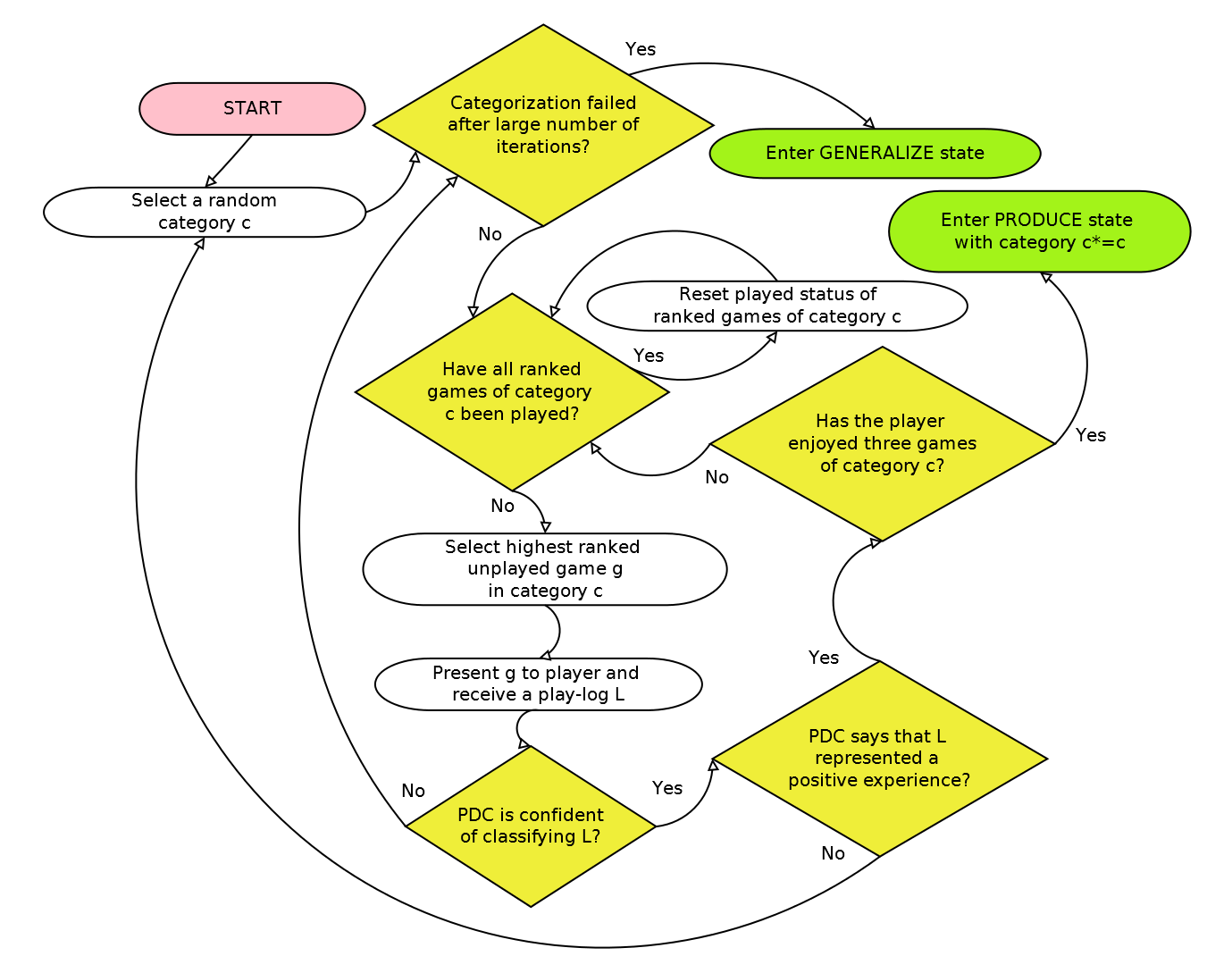}
\vspace*{-2mm}
\caption{The CATEGORIZE state in the IP model.}
\ec
\vspace*{-5mm}
\end{figure}

The goal of the IP model is three-fold: (a) discovering a target player's IP, (b) generating favorite content of variation for a target player whose IP has been determined as well as monitoring concept drift behavior, and (c) generalizing to a target player whose IP cannot be detected to avoid a system failure. We formulate the problem in the IP model as how to design decision-making procedures for all possible situations in personalized content generation at the on-line stage.
\par

 In general, our solution is strategically using all other models in the LBPCG framework to tackle various decision-making problems. We present our solution by a state machine as it provides a compact yet precise representation for our decision-making algorithms. Naturally, we establish three states in the IP model, i.e., CATEGORIZE, PRODUCE and GENERALIZE, to deal with three critical issues arising from the goal mentioned above, respectively.
\par

As illustrated in Fig. 2, the CATEGORIZE state is designed to detect a new target player's IP via games he/she has played. The main idea is as follows. For maximizing positive users experience and reliability in the IP detection, here we re-use selected games used in GPE learning. All the games are already grouped into game clusters with their content features and games in a cluster is ranked based on their $\gamma$ value achieved in GPE learning. Once a target player has played a game, the PDC model uses his/her play-log along with content features to predict his/her IP. If his/her behavior are consistent on games from a specific game category characterized by content features ${\bf c}^*$, it suggests that the target player enjoys games in this specific category. Then, the IP model leaves from the current state for the PRODUCE state.

\par
The PRODUCE state is shown in Fig. 3. It directs the IP model to use both the ICQ and the CC model to control the content generator to produce that quality content of variation in the specific game category determined in the CATEGORIZE state. Furthermore, the PDC model is employed to monitor whether a preference drift has taken place. Whenever a concept drift is detected, the IP model will get back into the CATEGORIZE state.

\par
To avoid a system failure, the GENERALIZE state shown in Fig. 4 is dedicated to
a crisis situation that a target player's IP cannot be determined after many attempts in the
CATEGORIZE state. Our idea in tacking this problem is exploiting the ICQ, the CC and the GPE models to generate quality games that is likely to be appreciated by public towards minimizing disappointment in playing randomly generated games of low quality. By using evaluation functions achieved in above three models, a target player receives only those games, ${\boldsymbol g} \in \mathcal{G}_{a\!c}$, of a large $\gamma$ value produced by $f({\boldsymbol g}, \Theta^*)$ learnt in the GPE model that indicates a high likelihood for such games to result in positive public experience.
Moreover, the PDC model is again employed to predict a target player's IP after a game was played. Once his/her IP is determined as is done in the CATEGORIZE state, the IP model would get into the PRODUCE state to generate content matching his/her IP.

\par
In summary, the state machine used in the IP mode provides an enabling technique geared towards generating personalized content for a target player whose preferences are initially unknown. Here we emphasize that our enabling techniques merely require play-logs of a target player in personalized content generation, which minimizes interruption to his/her experience.

\begin{figure}[t]
\label{fig:ip_produce}
\bc
\includegraphics[scale=0.28]{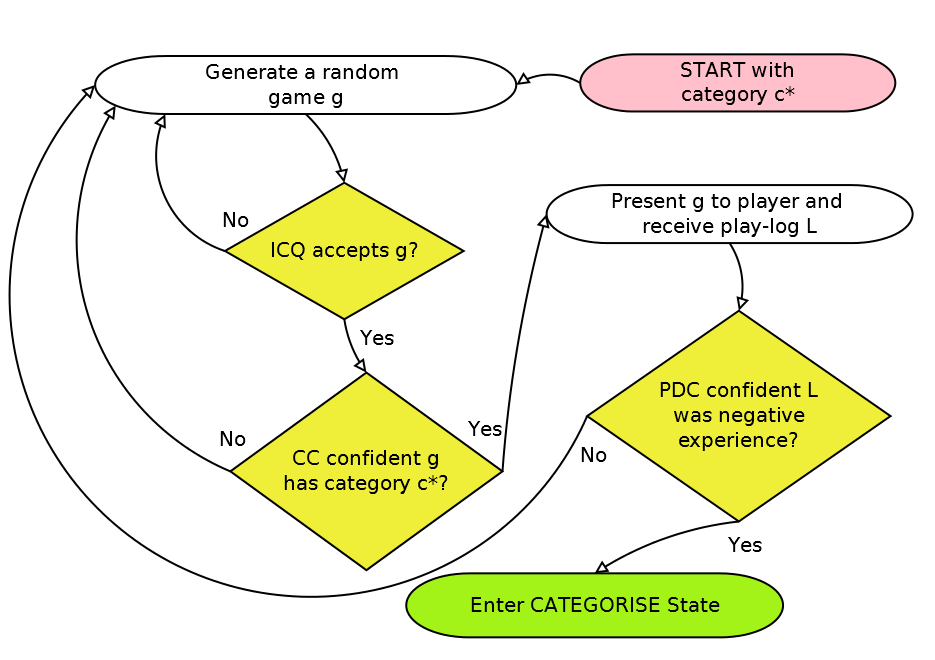}
\vspace*{-2mm}
\caption{The PRODUCE state in the IP model.}
\ec
\vspace*{-3mm}
\end{figure}

\begin{figure}[t]
\label{fig:ip_generalize}
\bc
\includegraphics[scale=0.28]{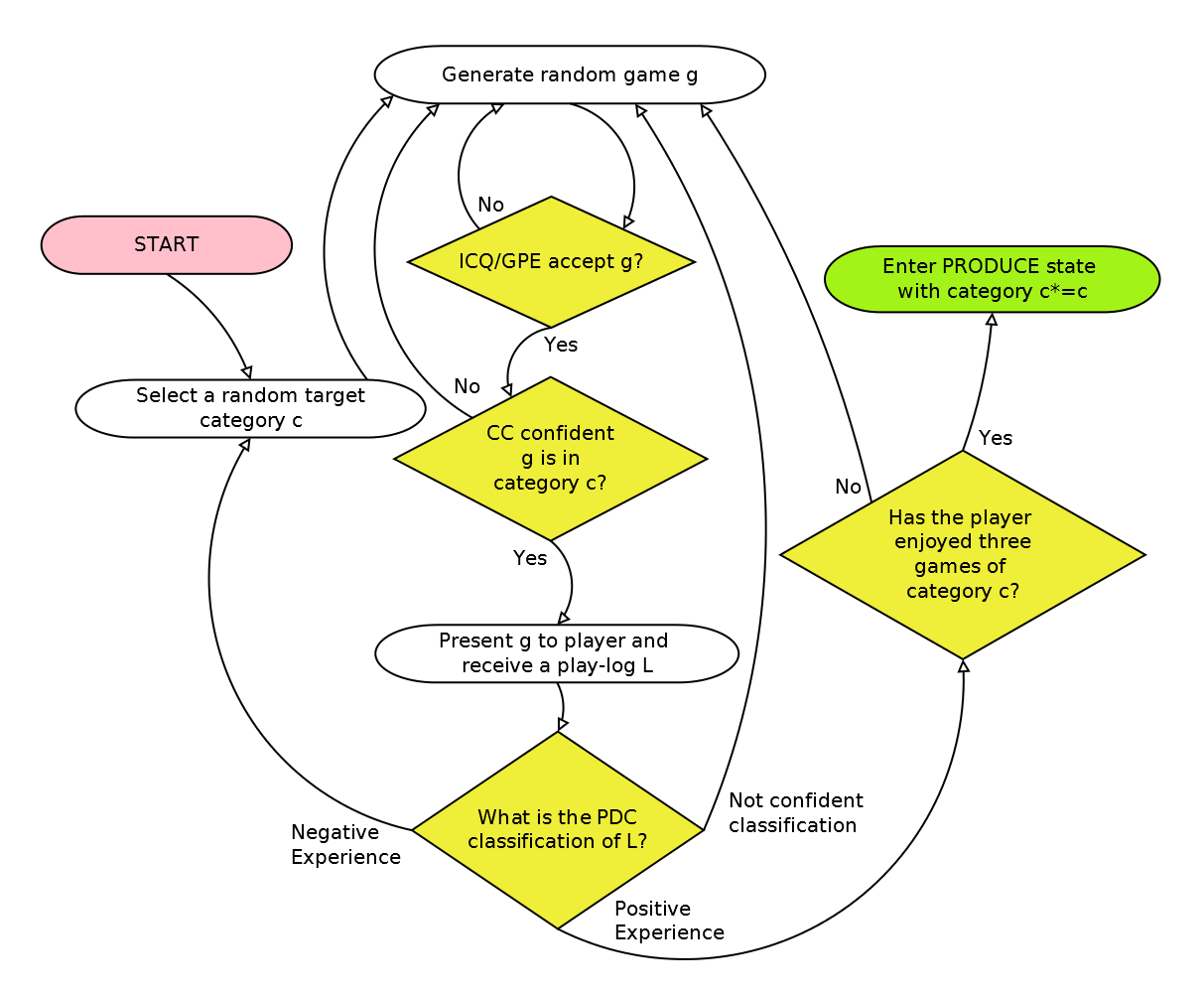}
\vspace*{-2mm}
\caption{The GENERALIZE state in the IP model.}
\ec
\vspace*{-5mm}
\end{figure}

%% file: tex/lbpcg_section_quake.tex
In this section, we present a proof-of-concept prototype by applying the LBPCG framework to the first-person shooter game, Quake \cite{Quake}, and report simulation results. To facilitate the presentation, hereinafter, we refer \emph{LBPCG-Quake} to the LBPCG framework applied to Quake and name its five composite models \emph{Quake-ICQ}, \emph{Quake-CC}, \emph{Quake-GPE}, \emph{Quake-PDC} and \emph{Quake-IP}.
\par

%% file: tex/lbpcg_section_quake_overview.tex
The first-person shooter Quake is a highly acclaimed game, which is open source, making it an ideal platform for research since we can modify it to suit our purposes. We have made various changes to it such as the ability to output play-logs.
OBLIGE~\cite{urlOblige} is a map generator for various Id Software games including Quake, which provides a command-line API that allows ordinal parameters to be set to control high-level attributes such as monster and weapon frequencies. OBLIGE is employed as the content generator and maps generated are the content in our LBPCG-Quake.
The majority of parameters effect aesthetic qualities of the level, making them somewhat less interesting. We have chosen a subset of nine OBLIGE parameters and in some cases combined them to form our content description vector. They are: skill level, overall monster count, amount of health packs, amount of ammo, weapon choice and four parameters consolidating the control of different types of monster. For parameters that we do not vary, we fixed their value to something sensible, e.g., the median value.

\par

With the subset of chosen parameters, the content vector space consists of a total of 116,640 games, which is certainly infeasible to exhaustively search by having the developers play each game. In addition, if one is to sample the content space randomly it becomes clear the majority of content has an extremely high level of difficulty. This poses an immediate problem for any PCG algorithms, as it would not be a good idea to present a nightmarishly difficult game to an experienced gamer, let alone a completely new player. There are also a lot of subtleties in the content parameters that need to be taken into account. For example, the overall number of monsters is controlled by one parameter, but there are four other parameters that control the proportion of different types of monsters that appear. In addition the play-logs we extract from Quake are high-dimensional. All of these issues make the problem of finding enjoyable content for a target player quite intricate and challenging. In overcoming these challenges, the LBPCG-Quake aims to produce maps that have just the right amount of monsters, power-ups and the best weapons for a target player.

\par

\begin{figure}
\centering
\includegraphics[scale=0.31]{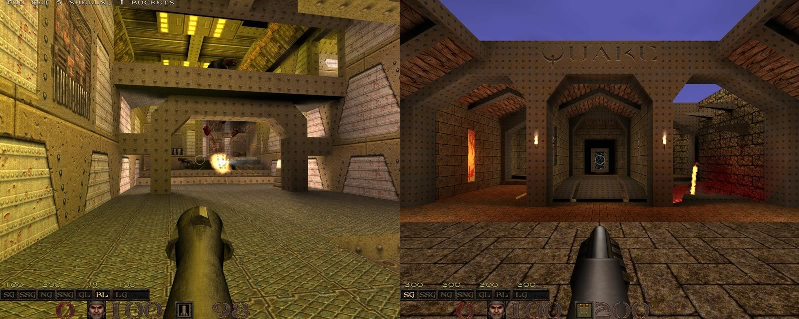}
\caption{The classic first-person shooter Quake.}
\label{fig:quake}
\vspace*{-5mm}
\end{figure}

%% file: tex/lbpcg_section_data_collection.tex
Both the Quake-ICQ and the Quake-CC were trained on a single expert who generated examples by playing and labeling games during the active learning. One can view this as a special case of developer training, where there is a single developer. The Quake-GPE and Quake-PDC, however, required that we put our framework out to the internet to simulate a scenario that gets as varied a cross-section of beta testers as possible.
To facilitate the public data collection, we developed a client/server architecture to collect surveys from people remotely.
For demonstration purposes, we have chosen only one content feature, \emph{difficulty}, consisting of five categories (Very Easy/Easy/Moderate/Hard/Very Hard) in the LBPCG-Quake. We believe that this is likely to be something that most people will have a preference for. To generate training data for the Quake-GPE learning, we choose 100 representative games to give to the beta testers; there are 20 games in each of five difficulty categories. The 100 games were selected via an active learning process. Using games from the Quake-CC training process is the ideal solution that in retrospect would have been done.
\par

The client was distributed via the web and advertised on websites such as Reddit, Facebook and the OBLIGE forums. In total 895 surveys were recorded from a total of 140 people. Each game was played roughly nine times each and players played on average approximately three to five games. However, one enthusiastic participant actually produced 154 surveys on their own. Our questionnaire consisted of two questions only: ``Did you enjoy the level?" (yes/no) and ``How do you rate it?" (Very Bad/Bad/Average/Good/Very Good).

\par

Approximately two thirds of surveys were positive feedback. Interestingly, the surveys showed that as game difficulty increases, the number of ``Very Good" labels also increases, but the ``Very Hard" category also has the most labels of ``Very Bad". The middle difficulties ``Easy" to ``Hard" have the least number of people labeling them as ``Very Bad". Additionally, as difficulty increases, less surveys are labeled as ``Average", indicating more polarized view points. Further analysis showed that ``Very Hard" and ``Very Easy" games cause the most disagreement amongst participants, whereas the middle difficulties caused less disagreement. This indicates that the games we gave to the beta testers were well selected since they caused controversy, potentially allowing us to distinguish between different types of players based on category. Nevertheless, such a survey at random is by no means guarantied to include different types of players.
\par

Whenever a game is played, our modified Quake engine produces a very large play-log consisting of 122 features. These features include statistics such as the average number of monsters killed per game tick, how much the mouse was moved in each direction and how many monsters of each type were killed. Every play-log received from the beta testers is linked to their answers to two questions mentioned above and the binary answer to the first question is used to train the Quake-PDC.

%% file: tex/lbpcg_section_lbpcg_quake_learning.tex
In the Quake-ICQ learning, we chose to use the $K$-medoids algorithm \cite{KMEDOIDS} for clustering analysis as it is insensible to initialization conditions. We set $K$ to 200 and chose 100 games from each cluster. Thus the clustering analysis was done on a reduced space of 20,000 games, a much more reasonable size to explore. As a result, the developer played and labeled 200 games to form the validation set $T_{ICQ}$. For active learning, a nonlinear support vector machine (SVM) with RBF kernel \cite{SVM} was used in Algorithm \ref{alg:icq_active} where it was initialized by two randomly chosen games in different classes. The active learning process was continued until the positive and negative error rates on $T_{ICQ}$ converged, indicating the model has similar performance with respect to identifying both acceptable and unacceptable games.

\par


In the Quake-CC learning, the Quake-ICQ model was first applied to find an acceptable subspace, $\mathcal{G}_a$,
and the validation set $T_{CC}$ was formed consisting of 110 labeled games with at least 20 games from each difficulty category. Since categorizing the difficulty of content is a multi-class classification task, we decomposed it into binary classification sub-tasks in active learning. As random forests (RF) \cite{Random_Forest} can identify useful features automatically in a complex task, we trained five RF binary classifiers separately with Algorithm \ref{alg:cc_active}, one for each difficulty category. Again,
 we used  the same initialization and stopping criteria as used in the Quake-ICQ active learning. After learning, we adopted the winner-take-all rule working on the confidence of five binary classifiers for decision-making.

\par

In the Quake-GPE learning, Algorithm \ref{alg:crowd_em} was directly applied to the data set collected in the public survey as described in Sect. \ref{subsect:quake_data_collection}.
After analysis, we found that the participant who made 154 surveys had inconsistent play-logs and feedback for the same games that he/she played for several times. Hence, we did not use his/her data in the Quake-GPE learning (as well as the Quake-PDC learning).
In addition, we employed a nonlinear support vector regressor (SVR) with the RBF kernel \cite{SVM} in Algorithm \ref{alg:crowd_em} instead of the logistic regressor suggested in \cite{Crowd_EM} as we faced a rather complex nonlinear regression problem. The Crowd-EM algorithm was terminated when the log-likelihood function (see \cite{Crowd_EM} for details) reached a local maximum after six EM epochs.

\par

 In the Quake-PDC learning, we employed multiple RF classifiers \cite{Random_Forest} to form an ensemble learner with Algorithm \ref{alg:pdc_ensemble}. In our simulations, we used four different thresholds, $0.0, 0.3, 0.6$ and $0.9$,  on $\alpha^{(p)}$ and $\beta^{(p)}$, respectively, and their combination resulted in 16 different training data sets. Accordingly, 16 RF classifiers were trained on them separately and then combined to form the ensemble learner used in the Quake-PDC model.

%% file: tex/lbpcg_section_simulation_result.tex
Based on the learning described in Sect. \ref{subsect:quake_learning}, we report simulation results on constituent models after the learning  and a preliminary evaluation on the LBPCG-Quake prototype.
\par

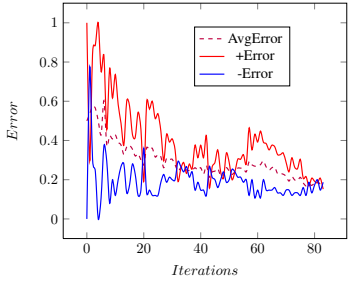
\begin{figure}[t]
\bc
\begin{tikzpicture}[scale=0.55]
\begin{axis}[xlabel=$Iterations$,ylabel=$Error$,legend style={at={(0.800000,0.900000)}}]
\addplot[smooth,purple,dashed,thick] plot coordinates {
(0.000,0.500)
(1.000,0.535)
(2.000,0.573)
(3.000,0.564)
(4.000,0.500)
(5.000,0.429)
(6.000,0.603)
(7.000,0.375)
(8.000,0.424)
(9.000,0.408)
(10.000,0.428)
(11.000,0.392)
(12.000,0.377)
(13.000,0.328)
(14.000,0.367)
(15.000,0.375)
(16.000,0.356)
(17.000,0.321)
(18.000,0.332)
(19.000,0.301)
(20.000,0.276)
(21.000,0.363)
(22.000,0.353)
(23.000,0.360)
(24.000,0.316)
(25.000,0.328)
(26.000,0.317)
(27.000,0.261)
(28.000,0.309)
(29.000,0.303)
(30.000,0.304)
(31.000,0.276)
(32.000,0.243)
(33.000,0.261)
(34.000,0.264)
(35.000,0.249)
(36.000,0.247)
(37.000,0.248)
(38.000,0.261)
(39.000,0.249)
(40.000,0.261)
(41.000,0.229)
(42.000,0.272)
(43.000,0.213)
(44.000,0.239)
(45.000,0.244)
(46.000,0.256)
(47.000,0.220)
(48.000,0.224)
(49.000,0.261)
(50.000,0.244)
(51.000,0.227)
(52.000,0.240)
(53.000,0.259)
(54.000,0.259)
(55.000,0.257)
(56.000,0.240)
(57.000,0.292)
(58.000,0.276)
(59.000,0.269)
(60.000,0.269)
(61.000,0.277)
(62.000,0.296)
(63.000,0.265)
(64.000,0.249)
(65.000,0.261)
(66.000,0.264)
(67.000,0.231)
(68.000,0.231)
(69.000,0.228)
(70.000,0.217)
(71.000,0.193)
(72.000,0.216)
(73.000,0.199)
(74.000,0.195)
(75.000,0.212)
(76.000,0.184)
(77.000,0.168)
(78.000,0.177)
(79.000,0.171)
(80.000,0.183)
(81.000,0.192)
(82.000,0.177)
(83.000,0.169)
};
\addlegendentry{AvgError}
\addplot[smooth,red,thick] plot coordinates {
(0.000,1.000)
(1.000,0.296)
(2.000,0.840)
(3.000,0.888)
(4.000,1.000)
(5.000,0.752)
(6.000,0.832)
(7.000,0.456)
(8.000,0.768)
(9.000,0.616)
(10.000,0.736)
(11.000,0.584)
(12.000,0.488)
(13.000,0.376)
(14.000,0.600)
(15.000,0.576)
(16.000,0.432)
(17.000,0.416)
(18.000,0.544)
(19.000,0.456)
(20.000,0.192)
(21.000,0.592)
(22.000,0.560)
(23.000,0.600)
(24.000,0.512)
(25.000,0.536)
(26.000,0.448)
(27.000,0.296)
(28.000,0.432)
(29.000,0.392)
(30.000,0.408)
(31.000,0.352)
(32.000,0.192)
(33.000,0.256)
(34.000,0.288)
(35.000,0.232)
(36.000,0.280)
(37.000,0.256)
(38.000,0.376)
(39.000,0.312)
(40.000,0.376)
(41.000,0.312)
(42.000,0.424)
(43.000,0.160)
(44.000,0.264)
(45.000,0.288)
(46.000,0.352)
(47.000,0.240)
(48.000,0.208)
(49.000,0.336)
(50.000,0.288)
(51.000,0.200)
(52.000,0.240)
(53.000,0.304)
(54.000,0.304)
(55.000,0.328)
(56.000,0.280)
(57.000,0.464)
(58.000,0.352)
(59.000,0.432)
(60.000,0.392)
(61.000,0.448)
(62.000,0.392)
(63.000,0.384)
(64.000,0.352)
(65.000,0.376)
(66.000,0.368)
(67.000,0.328)
(68.000,0.328)
(69.000,0.336)
(70.000,0.288)
(71.000,0.240)
(72.000,0.312)
(73.000,0.264)
(74.000,0.256)
(75.000,0.304)
(76.000,0.208)
(77.000,0.216)
(78.000,0.168)
(79.000,0.208)
(80.000,0.192)
(81.000,0.184)
(82.000,0.208)
(83.000,0.152)
};
\addlegendentry{+Error}
\addplot[smooth,blue,thick] plot coordinates {
(0.000,0.000)
(1.000,0.773)
(2.000,0.307)
(3.000,0.240)
(4.000,0.000)
(5.000,0.107)
(6.000,0.373)
(7.000,0.293)
(8.000,0.080)
(9.000,0.200)
(10.000,0.120)
(11.000,0.200)
(12.000,0.267)
(13.000,0.280)
(14.000,0.133)
(15.000,0.173)
(16.000,0.280)
(17.000,0.227)
(18.000,0.120)
(19.000,0.147)
(20.000,0.360)
(21.000,0.133)
(22.000,0.147)
(23.000,0.120)
(24.000,0.120)
(25.000,0.120)
(26.000,0.187)
(27.000,0.227)
(28.000,0.187)
(29.000,0.213)
(30.000,0.200)
(31.000,0.200)
(32.000,0.293)
(33.000,0.267)
(34.000,0.240)
(35.000,0.267)
(36.000,0.213)
(37.000,0.240)
(38.000,0.147)
(39.000,0.187)
(40.000,0.147)
(41.000,0.147)
(42.000,0.120)
(43.000,0.267)
(44.000,0.213)
(45.000,0.200)
(46.000,0.160)
(47.000,0.200)
(48.000,0.240)
(49.000,0.187)
(50.000,0.200)
(51.000,0.253)
(52.000,0.240)
(53.000,0.213)
(54.000,0.213)
(55.000,0.187)
(56.000,0.200)
(57.000,0.120)
(58.000,0.200)
(59.000,0.107)
(60.000,0.147)
(61.000,0.107)
(62.000,0.200)
(63.000,0.147)
(64.000,0.147)
(65.000,0.147)
(66.000,0.160)
(67.000,0.133)
(68.000,0.133)
(69.000,0.120)
(70.000,0.147)
(71.000,0.147)
(72.000,0.120)
(73.000,0.133)
(74.000,0.133)
(75.000,0.120)
(76.000,0.160)
(77.000,0.120)
(78.000,0.187)
(79.000,0.133)
(80.000,0.173)
(81.000,0.200)
(82.000,0.147)
(83.000,0.187)
};
\addlegendentry{-Error}
\end{axis}
\end{tikzpicture}
\vspace*{-3mm}
\caption{Test errors in the ICQ active learning process.}
\label{fig:icq_alplot}
\ec
\end{figure}

Fig. \ref{fig:icq_alplot} illustrates test error rates on $T_{ICQ}$ as the ICQ active learning process progresses. ``+Error" and ``-Error" stand for errors on positive and negative samples in $T_{ICQ}$, respectively. Also we report the \emph{half-time error rate} (HTER) defined as the average of ``+Error" and ``-Error" and denoted by ``AvgError". It is observed from Fig. \ref{fig:icq_alplot} that the overall error gradually decreases as more and more informative examples are presented and both positive and negative errors converge to an acceptable \emph{equal error rate} (EER), approximately $19\%$, after 81 iterations. Such a performance on $T_{ICQ}$ reflecting the distribution of Quake content space  demonstrates the effectiveness of our proposed active learning algorithm.
\par

\begin{figure}[t]
\bc
\begin{tikzpicture}[scale=0.55]
\begin{axis}[xlabel=$Iteration$,ylabel=$Error$,legend style={at={(0.900000,0.900000)}}]
\addplot[smooth,purple,thick] plot coordinates {
(1.000,0.800)
(2.000,0.800)
(3.000,0.636)
(4.000,0.650)
(5.000,0.543)
(6.000,0.510)
(7.000,0.570)
(8.000,0.527)
(9.000,0.497)
(10.000,0.475)
(11.000,0.450)
(12.000,0.467)
(13.000,0.398)
(14.000,0.377)
(15.000,0.386)
(16.000,0.428)
(17.000,0.410)
(18.000,0.412)
(19.000,0.419)
(20.000,0.394)
(21.000,0.394)
(22.000,0.373)
(23.000,0.273)
(24.000,0.387)
(25.000,0.396)
(26.000,0.344)
(27.000,0.321)
(28.000,0.338)
(29.000,0.336)
(30.000,0.313)
(31.000,0.320)
(32.000,0.302)
(33.000,0.313)
(34.000,0.273)
(35.000,0.259)
(36.000,0.256)
(37.000,0.265)
(38.000,0.278)
(39.000,0.219)
(40.000,0.235)
(41.000,0.264)
(42.000,0.229)
(43.000,0.237)
(44.000,0.243)
(45.000,0.257)
(46.000,0.265)
(47.000,0.247)
(48.000,0.248)
(49.000,0.255)
(50.000,0.247)
(51.000,0.247)
(52.000,0.257)
(53.000,0.247)
(54.000,0.264)
(55.000,0.257)
(56.000,0.264)
(57.000,0.255)
(58.000,0.255)
(59.000,0.275)
(60.000,0.245)
(61.000,0.254)
(62.000,0.256)
(63.000,0.244)
(64.000,0.296)
(65.000,0.253)
(66.000,0.253)
(67.000,0.273)
(68.000,0.245)
(69.000,0.261)
(70.000,0.263)
(71.000,0.255)
(72.000,0.247)
(73.000,0.236)
(74.000,0.253)
(75.000,0.236)
(76.000,0.246)
(77.000,0.238)
(78.000,0.265)
(79.000,0.283)
(80.000,0.262)
(81.000,0.272)
(82.000,0.344)
(83.000,0.302)
(84.000,0.295)
(85.000,0.321)
(86.000,0.308)
(87.000,0.307)
(88.000,0.301)
(89.000,0.337)
(90.000,0.313)
(91.000,0.292)
(92.000,0.293)
(93.000,0.263)
(94.000,0.281)
(95.000,0.284)
(96.000,0.264)
(97.000,0.288)
(98.000,0.296)
(99.000,0.288)
(100.000,0.254)
(101.000,0.271)
(102.000,0.290)
(103.000,0.272)
(104.000,0.289)
(105.000,0.272)
(106.000,0.263)
(107.000,0.254)
(108.000,0.280)
(109.000,0.289)
(110.000,0.281)
(111.000,0.283)
(112.000,0.288)
(113.000,0.271)
(114.000,0.287)
(115.000,0.299)
(116.000,0.262)
(117.000,0.290)
(118.000,0.286)
(119.000,0.312)
(120.000,0.304)
(121.000,0.296)
(122.000,0.288)
(123.000,0.303)
(124.000,0.296)
(125.000,0.297)
(126.000,0.314)
(127.000,0.297)
(128.000,0.295)
(129.000,0.304)
(130.000,0.305)
(131.000,0.314)
(132.000,0.314)
(133.000,0.296)
(134.000,0.312)
(135.000,0.295)
(136.000,0.287)
(137.000,0.287)
(138.000,0.297)
(139.000,0.289)
(140.000,0.288)
(141.000,0.295)
(142.000,0.287)
(143.000,0.285)
(144.000,0.287)
(145.000,0.287)
(146.000,0.269)
(147.000,0.296)
(148.000,0.288)
(149.000,0.269)
(150.000,0.278)
(151.000,0.288)
(152.000,0.270)
(153.000,0.287)
(154.000,0.288)
(155.000,0.290)
(156.000,0.279)
};
\addlegendentry{OverallError}
\end{axis}
\end{tikzpicture}
\vspace*{-3mm}
\caption{Test error in the CC active learning process.}
\label{fig:cc_learn_graph}
\vspace*{-5mm}
\ec
\end{figure}
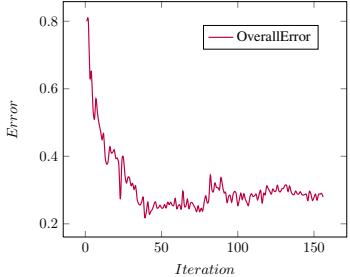

\par
 As the Quake-CC model classifies the game content into five categories,  we only report overall errors on $T_{CC}$ during the Quake-CC active learning in Fig. \ref{fig:cc_learn_graph} for clear visualization.  We observed that the model reached its optimal accuracy at 41 iterations and training beyond this leads to over-fitting to a specific category. At an early stop at iteration 41, the overall error  rate was $22\%$ and error rates for different difficulty categories were as follows: $17\%$  on ``Very Easy", $18\%$ on ``Easy", $35\%$ on ``Moderate", $25\%$ on ``Hard" and $20\%$ on ``Very Hard". A closer look at the confusion matrix revealed that almost all misclassification happened on two adjacent difficulty categories, e.g., ``Hard" is misclassified as ``Very Hard". Thus, such misclassification produces no catastrophic effect.  The use of rejection in decision making often increases accuracy and reliability
 \cite{duda01}.  However, it was found that applying the rejection in the Quake-CC model did not improve its performance considerably. Hence, we used this Quake-CC model of the optimal performance in the Quake-PDC and Quake-IP models to support the target player adaptation.
\par
As there is no ground-truth on popularity of beta test games and reliability of beta players, no test can be done for the Quake-GPE model. After the Quake-GPE learning, estimated popularity $\gamma$ for 100 beta test games, and reliability factors $(\alpha, \beta)$ of 139 beta players were directly used in the Quake-PDC learning and the Quake-IP model.

\par
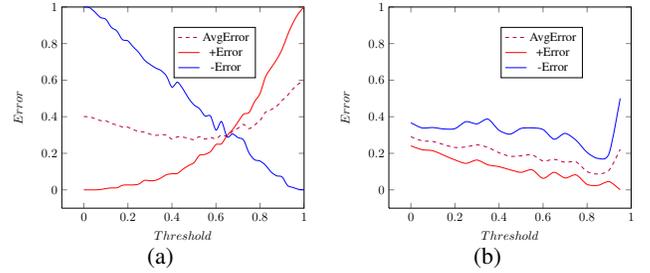
\begin{figure}[t]
\centering
\begin{minipage}[b]{0.45\linewidth}
\begin{tikzpicture}[scale=0.47]
\begin{axis}[xmax=1,ymax=1,xlabel=$Threshold$,ylabel=$Error$,legend style={at={(0.800000,0.900000)}}]
\addplot[smooth,purple,dashed,thick] plot coordinates {
(0.000,0.401)
(0.025,0.399)
(0.050,0.390)
(0.075,0.379)
(0.100,0.378)
(0.125,0.363)
(0.150,0.355)
(0.175,0.344)
(0.200,0.343)
(0.225,0.329)
(0.250,0.320)
(0.275,0.316)
(0.300,0.305)
(0.325,0.298)
(0.350,0.301)
(0.375,0.300)
(0.400,0.278)
(0.425,0.290)
(0.450,0.289)
(0.475,0.281)
(0.500,0.275)
(0.525,0.291)
(0.550,0.278)
(0.575,0.287)
(0.600,0.279)
(0.625,0.301)
(0.650,0.294)
(0.675,0.320)
(0.700,0.337)
(0.725,0.358)
(0.750,0.335)
(0.775,0.355)
(0.800,0.379)
(0.825,0.425)
(0.850,0.439)
(0.875,0.457)
(0.900,0.489)
(0.925,0.517)
(0.950,0.553)
(0.975,0.580)
(1.000,0.599)
};
\addlegendentry{AvgError}
\addplot[smooth,red,thick] plot coordinates {
(0.000,0.000)
(0.025,0.000)
(0.050,0.000)
(0.075,0.002)
(0.100,0.007)
(0.125,0.011)
(0.150,0.011)
(0.175,0.025)
(0.200,0.027)
(0.225,0.027)
(0.250,0.032)
(0.275,0.052)
(0.300,0.050)
(0.325,0.052)
(0.350,0.065)
(0.375,0.083)
(0.400,0.088)
(0.425,0.090)
(0.450,0.113)
(0.475,0.133)
(0.500,0.149)
(0.525,0.189)
(0.550,0.194)
(0.575,0.209)
(0.600,0.248)
(0.625,0.252)
(0.650,0.297)
(0.675,0.329)
(0.700,0.372)
(0.725,0.414)
(0.750,0.423)
(0.775,0.482)
(0.800,0.527)
(0.825,0.619)
(0.850,0.667)
(0.875,0.712)
(0.900,0.768)
(0.925,0.847)
(0.950,0.914)
(0.975,0.966)
(1.000,1.000)
};
\addlegendentry{+Error}
\addplot[smooth,blue,thick] plot coordinates {
(0.000,1.000)
(0.025,0.997)
(0.050,0.973)
(0.075,0.943)
(0.100,0.933)
(0.125,0.889)
(0.150,0.869)
(0.175,0.822)
(0.200,0.815)
(0.225,0.781)
(0.250,0.751)
(0.275,0.710)
(0.300,0.687)
(0.325,0.667)
(0.350,0.653)
(0.375,0.623)
(0.400,0.562)
(0.425,0.589)
(0.450,0.552)
(0.475,0.502)
(0.500,0.465)
(0.525,0.444)
(0.550,0.404)
(0.575,0.404)
(0.600,0.327)
(0.625,0.374)
(0.650,0.290)
(0.675,0.306)
(0.700,0.286)
(0.725,0.273)
(0.750,0.202)
(0.775,0.165)
(0.800,0.158)
(0.825,0.135)
(0.850,0.098)
(0.875,0.077)
(0.900,0.071)
(0.925,0.024)
(0.950,0.013)
(0.975,0.003)
(1.000,0.000)
};
\addlegendentry{-Error}
\end{axis}
\end{tikzpicture}
\vspace*{-7mm}
\bc
{\small (a)}
\ec
\end{minipage}
\hspace*{1mm}
\begin{minipage}[b]{0.45\linewidth}
\begin{tikzpicture}[scale=0.47]
\begin{axis}[xmax=1,ymax=1,xlabel=$Threshold$,ylabel=$Error$,legend style={at={(0.800000,0.900000)}}]
\addplot[smooth,purple,dashed,thick] plot coordinates {
(0.000,0.291)
(0.050,0.268)
(0.100,0.265)
(0.150,0.247)
(0.200,0.232)
(0.250,0.235)
(0.300,0.246)
(0.350,0.232)
(0.400,0.203)
(0.450,0.184)
(0.500,0.186)
(0.550,0.192)
(0.600,0.158)
(0.650,0.167)
(0.700,0.151)
(0.750,0.153)
(0.800,0.100)
(0.850,0.087)
(0.900,0.108)
(0.950,0.222)
};
\addlegendentry{AvgError}
\addplot[smooth,red,thick] plot coordinates {
(0.000,0.241)
(0.050,0.220)
(0.100,0.214)
(0.150,0.189)
(0.200,0.164)
(0.250,0.145)
(0.300,0.163)
(0.350,0.137)
(0.400,0.127)
(0.450,0.110)
(0.500,0.096)
(0.550,0.110)
(0.600,0.063)
(0.650,0.096)
(0.700,0.065)
(0.750,0.082)
(0.800,0.030)
(0.850,0.025)
(0.900,0.045)
(0.950,0.000)
};
\addlegendentry{+Error}
\addplot[smooth,blue,thick] plot coordinates {
(0.000,0.367)
(0.050,0.339)
(0.100,0.341)
(0.150,0.333)
(0.200,0.335)
(0.250,0.373)
(0.300,0.361)
(0.350,0.387)
(0.400,0.326)
(0.450,0.305)
(0.500,0.336)
(0.550,0.339)
(0.600,0.330)
(0.650,0.277)
(0.700,0.309)
(0.750,0.271)
(0.800,0.205)
(0.850,0.172)
(0.900,0.200)
(0.950,0.500)
};
\addlegendentry{-Error}
\end{axis}
\end{tikzpicture}
\vspace*{-7mm}
\bc
{\small (b)}
\ec
\end{minipage}
\caption{Test errors of the Quake-PDC models. (a) Error rates vs. confidence thresholds. (b)
Error rates vs. rejection thresholds.}
\label{fig:pdc_graphs}
\vspace*{-5mm}
\end{figure}

To test the Quake-PDC model, we used 10-fold cross-validation on all the play-logs
used in the Quake-GPE learning. As the RF provides a confidence value ranging from 0 to 1
on each decision, we further exploited such information to improve the Quake-PDC performance. By adjusting the decision boundaries with different confidence thresholds, we achieved different test results.
It is immediately apparent from Fig. \ref{fig:pdc_graphs}(a) that the positive/negative error rates converge to the EER of $29\%$ at a confidence threshold of 0.61. Furthermore, we applied rejection to increase the accuracy. As depicted in Fig. \ref{fig:pdc_graphs}(b), using a rejection threshold of 0.25 an HTER of 24\% was achieved, where approximately 25\% and 27\% of samples from the positive and negative classes, respectively, were rejected. Given the nature of noisy public surveys, we believe that overall performance is reasonable. Here, we emphasize that rejecting uncertain play-logs may avoid some catastrophic failure in the Quake-IP models. In our simulations, we used two aforementioned confidence and rejection thresholds in the Quake-PDC model for decision-making.

\par

 Ideally, testing the Quake-IP model requires a very extensive survey where a large number of reliable target players of different types are involved and play many games\footnote{Our LBPCG-Quake prototype is online available for public at \url{http://staff.cs.manchester.ac.uk/~kechen/lbpcg_quake}.}.
 In our preliminary evaluation,
 we managed to get four reliable people to simulate target players.  One player was completely inexperienced with video games, another was an expert player and the two remaining where roughly somewhere between these extremes. Apart from games generated by the Quake-LBPCG, each target player was also asked to play games generated by two comparison models: (a) \emph{ Balanced Model}, which presents equal numbers of games of each difficulty from the 100 beta test games and (b) \emph{Random Model}, which presents completely random games with no filtering whatsoever. Each player played 30 games, with 10 generated by each model. After playing each game, the player was asked the same question in the public survey, i.e., ``Did you enjoy the level?", and the feedback (yes/no) was recorded.
 \par
 To evaluate the performance of three different models, we define a scoring metric based on the feedback given by four target players.  For 30 games played by player $p$, let $N_{p,c}$ and $N^E_{p,c}$ denote the number of the games of difficulty category $c$ played and the  number of games of  difficulty category $c$ enjoyed by him/her (i.e., the number of positive feedback given by player $p$ to games of $c$), respectively.  Hence, the preference rate of player $p$ for games of difficulty category $c$ is $\rho_{p,c}= N^E_{p,c}/N_{p,c}$. For player $p$, let $N_{m,p,c}$ be the number of games of  difficulty category $c$ generated by model $m$. For model $m$ and player $p$, our scoring metric $S_{m,p}$ is defined as
 $$
 S_{m,p} = \sum_{c \in \mathcal{C}} \rho_{p,c} N_{m,p,c},
 $$
 where $\mathcal{C}$ is the set of five content categories in the LBPCG-Quake. Intuitively, a higher score awarded to a model indicates that the model produces more games of the difficulty category enjoyed by the player. Thus, we would use such a metric to measure the success of a model in terms of personalization.

 \par

\begin{figure}[t]
\centering
\includegraphics[scale=0.3]{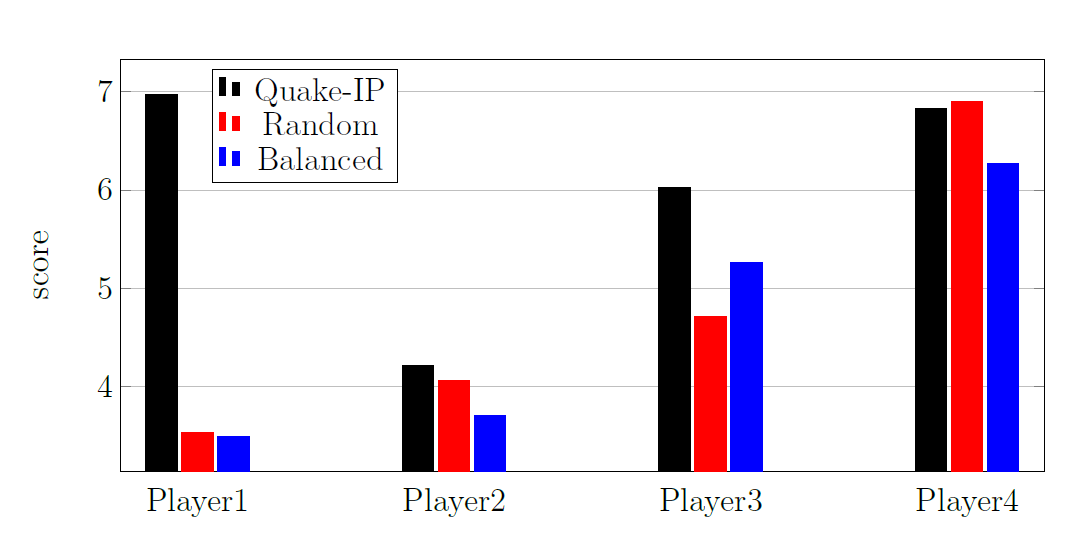}
\caption{Scores awarded to three different models vs. target players.}
\label{fig:quake}
\vspace*{-5mm}
\end{figure}

 Fig. \ref{fig:quake} depicts the scores awarded to different models bases on the feedback of four target players. It is evident from Fig. \ref{fig:quake} that the Quake-IP model clearly performs best for Players 1 and 3 as was awarded much higher scores than other two models. For Player 2, the Quake-IP model was awarded the highest score among three models but not by a large margin. For Player 4, the score awarded to the Quake-IP model was superior to the Balanced model but inferior to the Random model. Based on his feedback, it seemed that Player 4 had
 a particular preference for ``Easy" and ``Moderate" games. However, our analysis suggested that behavior (play-logs) in playing the games of his preference varied significantly from those in public survey. As a consequence, the Quake-CC model had little confidence in identifying his preference and this player never left the CATEGORIZE state (c.f. Fig. 2), which indicates the weakness of our random public survey.
 \par
 In summary, the LBPCG-Quake prototype carries out a proof of concept for our proposed LBPCG framework. Simulation results suggest that the LBPCG-Quake prototype is promising towards generating the personalized content.

%% file: tex/lbpcg_section_discussion.tex
In this section, we discuss issues arising from our LBPCG framework and relate it to previous work.
\par

As the LBPCG is a novel PCG framework, we would first elucidate its nature in light of the existing PCG taxonomy \cite{sbpcg}. First of all, our framework is neither on-line nor off-line in terms of the content generation. In our framework, the IP model via other models controls the content generator that presents new randomly generated content rather than solidified content for a target player at the on-line stage. On the other hand, our framework cannot be classified as ``on-line" as it does not create new content while each game is played but generates new content after each game is played. Hence, such a framework is of mixing on-line and off-line properties. Next, our framework adopts a typical generate-and-test style to generate content, which provides safe-guards for necessary content via the ICQ model but does not exclude optional content as long as is acceptable. Finally, our framework works on the generic random seed-parameter vector content representation, which leads to certain variation in generated content between different runs with identical parameters and hence has a stochastic generation nature.
\par

In the LBPCG, two models built up in the development stage lead to two content evaluation functions;
the ICQ model for measuring the acceptability of generated content and the CC model for extracting pre-defined content features of a legitimate game. According to the taxonomy in
\cite{sbpcg}, both models are direct evaluation functions. Moreover, two models implicitly encode developers knowledge in a data-driven way, which can be viewed as a mixture of theory-driven and data-driven functions
\cite{sbpcg}. In contrast, two models created in the public test stage lead to two novel content evaluation functions in terms of beta players' behavior/experience. By a crowd-sourcing learning, the GPE model yields a direct
evaluation function by mapping game content to public experience. The PDC model results in a function mapping players' behavior on specific content to their experience, which could be viewed as an ``interactive" evaluation
function \cite{sbpcg} despite its mixing on-line and off-line nature. Finally, the IP model in the on-line stage provides a new personalized PCG algorithm.
\par

In general, the personalization carried out in the LBPCG framework is based on the assumptions that  players of a certain type/style have similar playing behavior on the same games played and players whose playing behavior on a specific game is similar share the similar experience.
To generate quality yet personalized content for a target player, the LBPCG framework undergoes a series of tests by evaluating content in different models. Thus, the LBPCG framework may be limited by poor performance of the evaluation functions and the generator controlling procedure.
The enabling techniques proposed in this paper simply provide a tentative technical support to the LBPCG framework but are by no means sophisticated even though the LBPCG-Quake prototype demonstrates  usefulness.
We propose active learning algorithms in the ICQ and the CC models but there is no proof that such algorithms always converge and lead to satisfactory performance with a limited number of annotated examples.
 Apart from exploration of unsupervised learning, it is possible to employ artificial agents to facilitate the annotation to reduce the developers load \cite{Togelius2007}. Instead of playing all annotated games by themselves, developers may annotate content based on agents behavior/performance along with a small number of games they have to play. It is anticipated that such evaluation functions of both direct and simulation-based nature can be established at low cost but more robust as content space is explored thoroughly.
The performance of the GPE and the PDC models is critically determined by the quality of data collected in the public test stage. In our LBPCG-Quake simulation, we had no control at all on random beta players. However, beta testers of different types/styles may be recruited in a commercial development. Thus, it is likely to gain the sufficient information on players' behavior/experience if each representative game is played by enough beta testers of different types/styles and each beta tester plays at least a representative game in each content category.
In the on-line stage, target players may have to play a quite number of games before their preferred content categories are found. Hence, the number of content categories characterized by content features and the order in presenting games to unknown target players could affect their experience although such factors have been considered in our algorithms. As a result, content features should be pre-defined or selected carefully and more efficient algorithms in the IP model will be studied to lead to better playing experience.
\par

SBPCG \cite{sbpcg} is a generic PCG framework and dominates the current PCG development. As a core technique, evolutionary computation has been widely used in SBPCG. While both SBPCG and our LBPCG share the generate-and-test style, there are several main differences between them as follows. First, content representation in SBPCG is a critical issue that has to be dealt with differently for different content types. In contrast, our LBPCG works on a universal content parameter representation regardless of content types. Next, reducing the content search space is carried out differently in SBPCG and the LBPCG; our LBPCG employs a number of data-driven evaluation functions that encode developers knowledge implicitly and model beta players' behavior/experience, while SBPCG often relies on evaluation functions constructed with a static heuristic based on explicit prior knowledge and target players' behavior/experience. Last, perhaps the most different, our LBPCG adopts an alternative fashion in generating the game content; the content generation is driven by target players' historical behavior and safeguarded by different evaluation functions for catastrophic failure avoidance and quality assurance. In contrast, SBPCG uses either a heuristic search or population-based solutions \cite{sbpcg}.
\par

While our LBPCG framework presents an alternative solution to PCG, it is possible to combine the LBPCG with existing PCG techniques. Evaluation functions learnt in the ICQ and the CC models may be used in SBPCG  for catastrophic failure avoidance, quality assurance and content space organization. While we advocate the use of learning-based enabling techniques, the ICQ and CC models are replaceable by any hand-crafted evaluation functions developed with prior knowledge \cite{sbpcg} for the same purpose as long as such evaluation functions of high performance have been developed in advance. From the perspective of personalization, our LBPCG is yet another experience-driven PCG approach \cite{experience11}. Two novel evaluation functions achieved in the GPE and the PDC models could be integrated into experience-driven SBPCG  \cite{experience11} to facilitate limiting the search space. In addition, we adopt a subjective play experience modeling (PEM) approach in our enabling techniques. If appropriate apparatus is already deployed into a platform, objective PEM approaches \cite{experience11}
may be easily used in the GPE and PDC models to substitute or enhance the current enabling techniques. \par

In conclusion, we have presented the LBPCG framework and enabling techniques in order to respond to several challenges in existing PCG techniques \cite{sbpcg}, including content representation, catastrophic failure avoidance, content generation speed-up, incorporating player models into evaluation functions and combination of theory-driven and interactive evaluation functions. Our framework is demonstrated via a proof-of-concept prototype based on Quake, which leads to promising results. In our ongoing researches, we will be further developing the LBPCG to overcome its limitation and exploring state-of-the-art machine learning and other relevant approaches to improve its enabling techniques.

%% file: tex/lbpcg_section_acknow.tex
Authors are grateful to all anonymous public players for their feedback in the LBPCG-Quake public survey and four target players who participated in our LBPCG-Quake simulations. K. Chen is the corresponding author.